\documentclass[10pt,twocolumn,letterpaper]{article}

\usepackage[pagenumbers]{cvpr} 

\usepackage{graphicx}
\usepackage{amsmath}
\usepackage{amssymb}
\usepackage{booktabs}

\usepackage[pagebackref,breaklinks,colorlinks]{hyperref}

\usepackage[capitalize]{cleveref}
\crefname{section}{Sec.}{Secs.}
\Crefname{section}{Section}{Sections}
\Crefname{table}{Table}{Tables}
\crefname{table}{Tab.}{Tabs.}

\def\cvprPaperID{*****} 
\def\confName{CVPR}
\def\confYear{2022}

\begin{document}

\title{Deep Image Style Transfer from Freeform Text}

\author{Tejas Santanam, Mengyang Liu, Jiangyue Yu, Zhaodong Yang\\
Georgia Institute of Technology\\
{\tt\small tsantanam@gatech.edu, mliu444@gatech.edu, jyu478@gatech.edu, halyang@gatech.edu}
}
\maketitle

\begin{abstract}
   This paper creates a novel method of deep neural style transfer by generating style images from freeform user text input. The language model and style transfer model form a seamless pipeline that can create output images with similar losses and improved quality when compared to baseline style transfer methods. The language model returns a closely matching image given a style text and description input, which is then passed to the style transfer model with an input content image to create a final output. A proof-of-concept tool is also developed to integrate the models and demonstrate the effectiveness of deep image style transfer from freeform text.
\end{abstract}

\section{Introduction}
\label{sec:intro}

Style transfer allows for one to imagine artistic combinations that do not actually exist, such as \emph{The Last Supper} painted by Kandinsky, or \emph{Nighthawks} painted by Miro. The ability to gain experience in building a model for a similar task interested us greatly. However, we wanted to go beyond some of the generic style transfer models we had seen. A common issue we had noticed in style transfer was the lack of ability to clearly specify \emph{what} within a style we wanted to transfer. Thus, the paper uses the user's text input to extract desired style information from a user description. An NLP model will be used to return an image as the style image based on the user's text input of style description. We then propose a high-quality style transfer model that attempts to both apply the desired style to our target image, but also do so in a way that preserves the key features of the target image. This paper finds creative ways to measure performance - something that is not straightforward in style transfer models. We also propose methods to handle image inputs of different types and sizes from the text-image query pipeline we build to get images from user text descriptions. The successful execution of this method allows for anyone to utilize style transfer even without a given style image. This is helpful for people not well-versed in art, or can describe a style they like but do not know the names of any artworks of that type.

\subsection{Related Works}

Keyword extraction is one of the most popular topics in NLP. Beginning in the last century, traditional machine learning method like C4.5 decision tree \cite{c4.5} and Naive Bayes Classifier \cite{KEA} has been applied to keyword extraction.-
Recently, recurrent neural network (RNN) models are exploited in the keyword extraction domain. As an example, Zhang et al.\cite{keyphrase} proposed a joint-layer RNN model to overcome the limitation of short content in tweets. And for style transfer, inspired by the power of Convolutional Neural Networks (CNNs), Gatys et al. \cite{gatys2015neural}. first applied a CNN to reproduce famous painting art styles on ordinary images. They utilized CNNs to model the content of a photo and the style information of an artwork. Their experimental results demonstrated CNN's capability of extracting content information from an ordinary photo and style information from a well-known artwork. The main idea of their research is to iteratively optimize an image with the objective of matching desired CNN feature distributions, which involves both the photo’s content information and the artwork’s style information.\\

After Gatys's work, lots of follow-up studies have been conducted to either improve or extend the Neural Style Transfer (NST). By controlling perceptual factors, Gatys et al. themselves \cite{control} propose several slight modifications to improve their previous algorithm. DeepDream \cite{Deepdream} is the first attempt to produce artistic images by reversing CNN representations with image-optimization-based online image reconstruction techniques. Their basic idea is to first model and extract style and content information from the corresponding style and content images, recombine them as the target representation, and then iteratively reconstruct a stylized
result that matches the target representation. Non-parametric image-optimisation-based NST is built on the basis of Non-parametric Texture Modelling with MRFs. Li and Wand \cite{DBLP} are the first to propose an MRF-based NST algorithm. They model the style in a non-parametric way and introduce a new style loss function which includes a patch-based MRF prior. It performs especially well for photorealistic styles due to patch-based MRF loss. And even before the appearance of Neural Networks, there were studies of artistic rendering (AR) algorithms that enables style transfer without Neural Networks. Stroke-based rendering (SBR) \cite{SBR}, which incrementally composites strokes to match a source photo to produce non-photorealistic imagery, is one of the most famous AR algorithms.\\

In summary, existing style transfer models are all relying on actual image input as style information for the transfer model. The state-of-the-art methods of NST are still modifying the architecture of neural networks to optimize the extraction of style information from style image input. For this paper, we propose a style transfer pipeline that only requires a general text input as the style information, and an image input as the content information. In our pipeline, we use a text-image query engine that is able to generate style image output based on the text input. We borrow methods from SemArt\cite{semart}, a fine-art understanding model that maps paintings and artistic comments into a common semantic space, thus enabling comparison in terms of semantic similarity. We utilize the encoding methods of painting images and artistic description to match text input with candidate images that are closest to the encoding of the text in semantic space, to generate an image according to the description text users input. Then we build our style transfer model based on traditional style transfer metrics. Content loss and style loss from traditional methods\cite{gatys2015neural} are used to measure how well we transferred the style. Several experiments are conducted to demonstrate the performance of our pipeline, which is capable of generating a style-transferred image from text input and image input. To the best of our knowledge, our pipeline is the first to realize style transfer with only general text input as the style information.\\
\\

\section{Method and Technical Approach}
\label{sec:formatting}

\begin{figure}[t]
  \centering
   \includegraphics[width=1\linewidth]{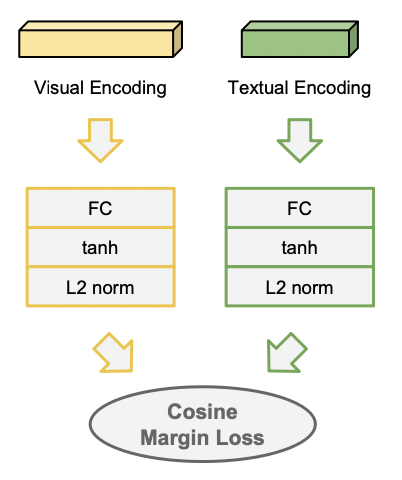}

   \caption{Fully Connected Neural Network for Mapping Textual and Visual Representations Into a Joint Embedding Space}
   \label{fig:CML}
\end{figure}

The technical approach consists of two primary parts: a text-image query engine and a style transfer model. After the user enters their free-form stylistic text description and stylistic title in text boxes, that text will be input into a text-to-image retrieval model, and a candidate image will be picked from the image bank based on the closest match to the input text. These images will then serve as the style images to generate the target embeddings in the style transfer step. To achieve this, we built an image retrieval algorithm based on the work of Garcia et al. \cite{semart}, where the basic idea is to first encode both the images and the texts into vector representations and then learn transformation functions that project these vectors into a joint embedding space so that images and comments from the same sample are mapped closer than images and comments from different samples in terms of the given similarity function $d$:
$$ d(\textit{\textbf{p}}_k^{text}, \textit{\textbf{p}}_k^{vis}) < d(\textit{\textbf{p}}_k^{text}, \textit{\textbf{p}}_j^{vis}), \forall k \neq j$$
and 
$$ d(\textit{\textbf{p}}_k^{text}, \textit{\textbf{p}}_k^{vis}) < d(\textit{\textbf{p}}_j^{text}, \textit{\textbf{p}}_k^{vis}), \forall k \neq j$$
where $\textit{\textbf{p}}^{text}$ and $\textit{\textbf{p}}^{vis}$ are the embeddings of text and image vectors after projection, respectively. The structure of this model is shown in Figure \ref{fig:CML}. After the images and texts are encoded into vectors, they are fed to different linear layers (denoted as FC in the figure) with activation layers and normalization layers on top to be mapped into embeddings with the same dimensionality. Here the loss function is calculated by combining cosine similarity with margin:
$$ L(\textit{\textbf{p}}_k^{text}, \textit{\textbf{p}}_j^{vis}) = \begin{cases}
  1-cos(\textit{\textbf{p}}_k^{text}, \textit{\textbf{p}}_j^{vis})  & \text{if } k = j \\
  max(0, cos(\textit{\textbf{p}}_k^{text}, \textit{\textbf{p}}_j^{vis})-m) & \text{if } k \neq j
\end{cases}$$
The encoding of the images and texts can be accomplished by multiple approaches. In the baseline model, ResNet \cite{resnet}, which improves the performance of traditional CNNs by using shortcut connections to connect the input of a layer to the output of a deeper layer, is used to generate the visual encodings. Depending on the number of layers, ResNet has various versions, such as ResNet50 and ResNet152. More specifically, We used the output of the last layer of ResNet50 as the visual encoding. For text encoding, the baseline model utilizes the bag-of-words model, where each artistic comment is encoded as a term frequency-inverse document frequency (tf-idf) vector by weighting each word in the comment by its relevance within the corpus (a comment vocabulary that contains all the alphabetic words that appear at least ten times in the training set). In addition, the titles of the images are also encoded as tf-idf vectors using a title vocabulary to provide extra textual information to the model. Besides this baseline model, we also experimented with a modified version of the model where we used a pre-trained BERT\cite{bert} model from \href{https://huggingface.co}{Hugging Face} instead of the bag-of-words model to encode the texts. Based on transformers, BERT could generate contextualized word embeddings based on the input, so we hoped that this modified model could outperform the baseline.\\

The image that output results from the NLP model is connected to the style transfer model as style image input. To measure how well the style transfer model performs quantitatively, we focus on two metrics: content loss and style loss. The content loss measures how much information we kept from our original image while the style loss measures how well we transferred the style. The losses are formulated as below,

\begin{align*}
    L_{content}(\mathbf{p}, \mathbf{x}, l) = \frac{1}{2}\sum_{i,j}(F_{ij}^l - P_{ij}^l)^2 \\
    E_l = \frac{1}{4N_L^2M_l^2}\sum_{i,j}(G_{ij}^l - A_{ij}^l)^2 \\
    L_{style}(\mathbf{a}, \mathbf{x}) = \sum_{l=0}^Lw_lE_l \\
\end{align*}
where $l$ is the layer, $\mathbf{p}$ and $\mathbf{x}$ are the generated image and original image. $F^l$ and $P^l$ are the feature representation output of layer $l$. $\mathbf{x}$ and $\mathbf{a}$ be the original image and the image that is generated and $G^l$ and $A^l$ are their respective style representations in layer $l$. We can also show the generated image to qualitatively show the performance of our framework.

We have a lot of choices for the general framework of the style transfer model from some of the existing literature \cite{gatys2015neural, johnson2016perceptual}. We use the model from Gatys et al. \cite{gatys2015neural} as our baseline model that we will try and compare against for improvements. We also focused on the architecture design of the transfer model, and tried different feature extractors like CNN, ResNet \cite{he2016deep} or ViT\cite{dosovitskiy2020image}. We can make different changes such as the number of layers, type of layers, kernel size, stride, number of filters, and more to see the impact on performance. Ultimately, the two best feature extractors were SqueezeNetv1.1 and ElasticNet\_v2.\\

For the development of the both the NLP and style transfer models, the PyTorch framework is utilized. For the style transfer model we implemented, the parameter sizes are shown in Table \ref{tab:parametersize}. Since we only need to use the model to extract the features from the style image and content image, the size of learnable parameters is actually the same as the size of the content image. 

\begin{table}
  \centering
  \begin{tabular}{cc}
    \toprule
    Model & Parameter Size \\
    \midrule
    SqueezeNetv1.1 & 1.2M \\
    EfficientNet\_v2 & 24M \\
    \bottomrule
  \end{tabular}
  \caption{Style Transfer Model Parameter Size}
  \label{tab:parametersize}
\end{table}

\section{Results}

\subsection{Dataset and Data Processing}
The source of our image dataset comes from WikiArt. WikiArt (\href{https://www.wikiart.org}{wikiart.org}) is an online art encyclopedia which contains more than 160,000 artworks from more than 3,000 artists classified according to styles, periods, and more. With this dataset, we are able to collect paintings of different styles that our models leverage to learn for style transfer tasks.\\

In paper, we also make use of a crawler developed by others to better retrieve the images from the online datasets. WikiArt Crawler (\href{https://github.com/asahi417/wikiart-image-dataset#wikiart-general}{GitHub link}) is a python library developed by Asahi Ushio to download and process images from WikiArt through WikiArt API, and it is used to retrieve the paintings of specific types and styles that we need for our tasks.\\

We also utilized the SemArt dataset created by Garcia et al. \cite{semart} for the text-to-image retrieval task. The painting images and their corresponding metadata within this dataset are collected from the Web Gallery of Art (\href{https://www.wga.hu}{WGA}), a website with more than 44,809 color images of European fine-art reproductions between the 8th and the 19th century. Furthermore, artistic comments are collected for each painting image, resulting in 21,384 total samples (over 20 GB of data) where each sample is a triplet of an image, an artistic comment and a number of attributes. The data are collected by Noa Garcia from the School of Informatics and Digital Engineering at Aston University, for studying semantic art understanding through images of fine-art paintings and artistic comments. Only images whose field form is set as painting are collected from artworks, as opposed to images of other forms of art such as sculpture or architecture. Images that are not associated with any comment are omitted, and irrelevant metadata fields, such as author’s birth and death and current location are removed as well.

\subsection{Baseline Evaluation}

\begin{table}
  \centering
  \begin{tabular}{l*{4}{c}r}
    \toprule
    Content Source, Style Source & Content Loss & Style Loss \\
    \midrule
    Tubingen, Composition VII & 9300.5 & 3223.1 \\
    Tubingen, Scream & 10547 & 3857 \\
    Tubingen, Starry Night & 13756.4 & 3006 \\
    Tubingen, Waterlilies & 17558.5 & 2795.3 \\
    Tubingen, Persistance of Memory & 16145.1 & 10464.7 \\
    Muse, Composition VII & 9078.5 & 4292.1 \\
    Muse, Scream & 13690.2 & 5955 \\
    Muse, Starry Night & 19967.6 & 6542.9 \\
    \bottomrule
  \end{tabular}
  \caption{Gatys et al. Baseline Results}
  \label{tab:baseline}
\end{table}

\begin{figure}[t]
  \centering
   \includegraphics[width=1\linewidth]{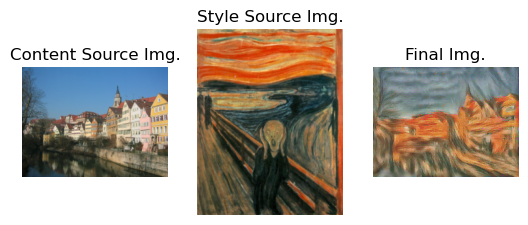}

   \caption{An Example of Baseline Model Style Transfer}
   \label{fig:baselinescream}
\end{figure}

\begin{figure}[t]
  \centering
   \includegraphics[width=1\linewidth]{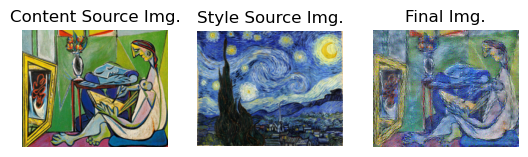}

   \caption{Another Example of Baseline Model Style Transfer}
   \label{fig:baselinestarry}
\end{figure}

The content and style losses from the baseline model on some test images can be seen in Table \ref{tab:baseline}. For this model, SqueezeNet was used as the feature extractor \cite{https://doi.org/10.48550/arxiv.1602.07360} and the model was as described in Gatys et al. \cite{gatys2015neural}.
A visual example of the baseline model output is shown in Figure \ref{fig:baselinescream} and Figure \ref{fig:baselinestarry}. While the model produces good visual results, the loss values are still quite high, and the inability of the model to take in style text input is something that the methodology in this paper improves upon.

\subsection{Text-to-Image Retrieval Results}
\begin{figure}[t]
  \centering
   \includegraphics[width=1\linewidth]{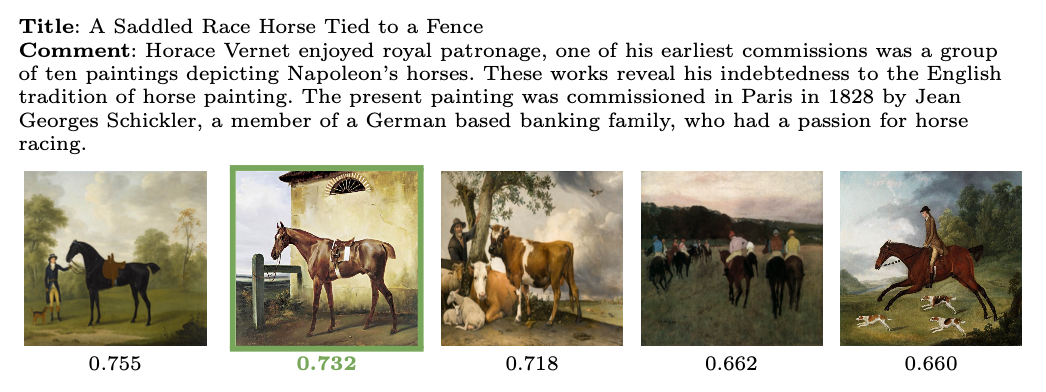}

   \caption{An Example of Text-to-Image Retrieval}
   \label{fig:text-to-image}
\end{figure}

We trained the baseline text-to-image retrieval model for 30 epochs. Other hyperparameters are set as shown in Table \ref{tab:hyperparametersNLP}. When given a text as input, painting images are ranked by the model according to their similarity to the given text, as shown in Figure \ref{fig:text-to-image}. The retrieval results are evaluated based on two metrics: median rank (MR) and recall rate at K (R@K), with K being 1, 5 and 10. MR is the median of the ranking positions of true images among the whole test set and the recall rate at K is the rate of samples for which the true image is within the top K ranked images. The evaluation results are shown in Table \ref{tab:CML} and the values are similar to, if not slightly better than, those reported by Garcia et al \cite{semart}.\\

We also trained the modified model with the same hyperparameter setting, and the results are shown in Table \ref{tab:CML_BERT}. To our surprise, the modified model which uses BERT for textual encoding performs much worse than the baseline model. There are several possible explanations. First, since the BERT model is pre-trained on a large corpus of English texts for general use, it may not be necessarily suitable for tasks in highly specific domains such as artistic comments. In the contrast, bag-of-words model might be able to better capture the linguistic characteristics in this task since the vocabulary is completely built from the corpus of artistic comments. Besides, we froze the parameters of the pre-trained BERT model during training and only trained the weights of the single linear layer on top of it in order to save time. This might also be a reason for the unsatisfactory performance.

\begin{table}
  \centering
  \begin{tabular}{p{0.8cm}p{1cm}p{0.8cm}p{1.4cm}p{1.1cm}}
    \toprule
    Epochs & Learning Rate & Batch Size & Embedding Size & Loss Margin \\
    \midrule
    30 & 0.001 & 28 & 128 & 0.1 \\
    \bottomrule
  \end{tabular}
  \caption{Text-to-Image Model Hyperparameters}
  \label{tab:hyperparametersNLP}
\end{table}

\begin{table}
  \centering
  \begin{tabular}{l*{5}{c}r}
    \toprule
     & MR & R@1 & R@5 & R@10 \\
    \midrule
    Validation Set & 9 & 0.184 & 0.408 & 0.543 \\
    Test Set & 9 & 0.177 & 0.399 & 0.531 \\
    \bottomrule
  \end{tabular}
  \caption{Baseline Text-to-Image Model Results}
  \label{tab:CML}
\end{table}

\begin{table}
  \centering
  \begin{tabular}{l*{5}{c}r}
    \toprule
     & MR & R@1 & R@5 & R@10 \\
    \midrule
    Validation Set & 145 & 0.010 & 0.051 & 0.096 \\
    Test Set & 153 & 0.009 & 0.047 & 0.093 \\
    \bottomrule
  \end{tabular}
  \caption{Modified Text-to-Image Model Results}
  \label{tab:CML_BERT}
\end{table}

\begin{figure}
    \centering
    \includegraphics[width=1\linewidth]{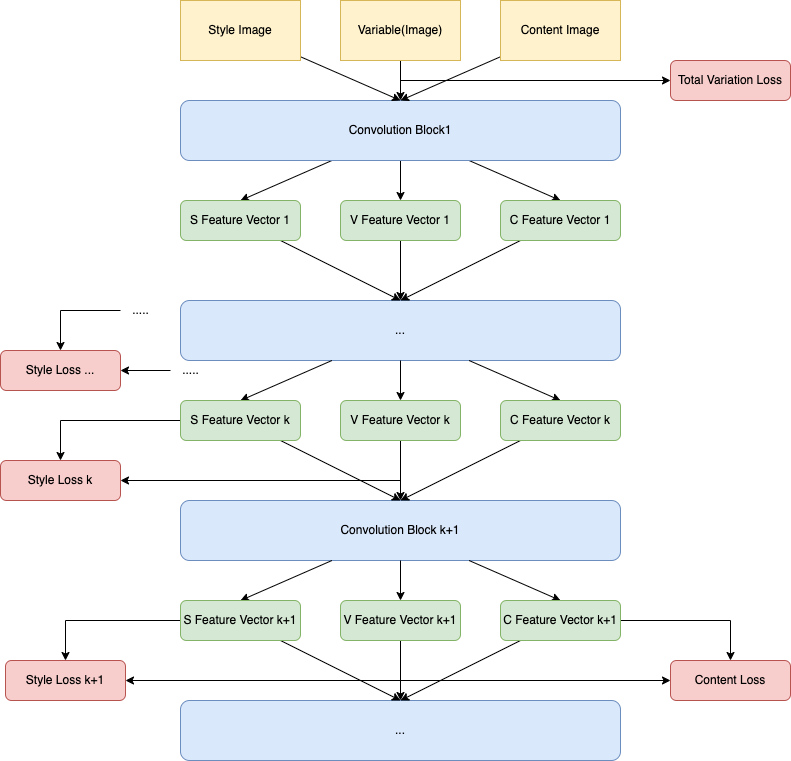}
    \caption{Style Transfer Framework Architecture}
    \label{fig:architecture}
\end{figure}

\begin{figure}[t]
  \centering
   \includegraphics[width=1\linewidth]{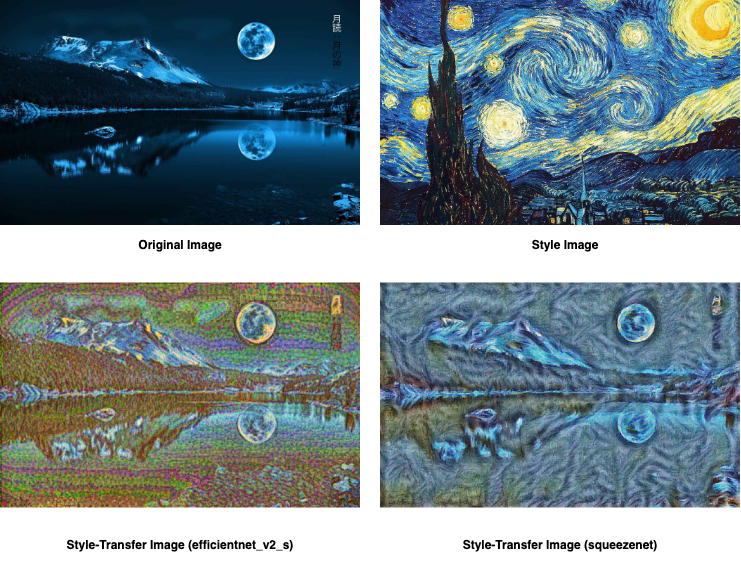}

   \caption{Examples of Style Transfer Output}
   \label{fig:styleexamples}
\end{figure}

\subsection{Style Transfer Results}
Our style transfer model is built on pre-trained foundational computer vision models such as EfficientNet\_v2 \cite{tan2019efficientnet}, SqueezeNet \cite{iandola2016squeezenet}, and more. These models are pre-trained on the massive ImageNet dataset for image classification. By computing the content loss, style loss, and total variation loss, we gradually update the synthesized image until we obtain an image with transferred style. Our style-transfer framework containing multiple convolutional blocks has the architecture shown in Figure \ref{fig:architecture}. After trying various feature extractors, the selected feature extractor for our final model was SqueezeNetv1.1, which has 2.4 times less computation and slightly fewer parameters than SqueezeNetv1.0, without sacrificing accuracy. The model using SqueezeNet was able to provide a style transferred final image given a content image and style image in under 15 seconds, compared with the next best feature extractor in EfficientNet\_v2 which took around 5 minutes. The model with EfficientNet\_v2 preserved content slightly better than SqueezeNet, but SqueezeNet also preserved style better from both a visual perspective and a style loss perspective, which was in line with our motivations. The final model uses an Adam optimizer with an initial learning rate of 3 and a decayed learning rate of 0.1 that decays after 180 iterations. Parameter tuning was also performed on content layers, content weight, style layers, style weight, and TV weight. All of the parameters were tuned to minimize content loss, style loss, and TV loss. In addition to tuning based on loss values, there was also a subjective component of quality of the output image. There were some cases where low loss values resulted in a poor image, and cases where decent images were generated when there were higher loss values. Ultimately, the final set of hyperparameters that most frequently balanced low loss and subjectively clear images can be seen in Table \ref{tab:hyperparametersST}. A results comparison of loss values for an example image pair using SqueezeNet and EfficientNet\_v2 can be seen in Table \ref{tab:results}. Some examples of model output can be seen in Figure \ref{fig:styleexamples}. After tuning, the style transfer model is able to consistently achieve values less than 1500, and often less than 1000, for all three loss types regardless of the input images. These results are much better than the numbers in Table \ref{tab:baseline}.

\begin{table}
  \centering
  \begin{tabular}{p{1cm}p{1cm}p{1cm}p{2cm}p{1cm}}
    \toprule
    Content Layers & Content Weight & Style Layers & Style Weight & TV Weight \\
    \midrule
    3 &0.001 & 2,4,6,7&400,50,10,5 & 0.005 \\
    \bottomrule
  \end{tabular}
  \caption{Style Transfer Model Tuned Hyperparameters}
  \label{tab:hyperparametersST}
\end{table}

\subsection{Proof-Of-Concept}
\begin{figure}[t]
  \centering
   \includegraphics[width=1\linewidth]{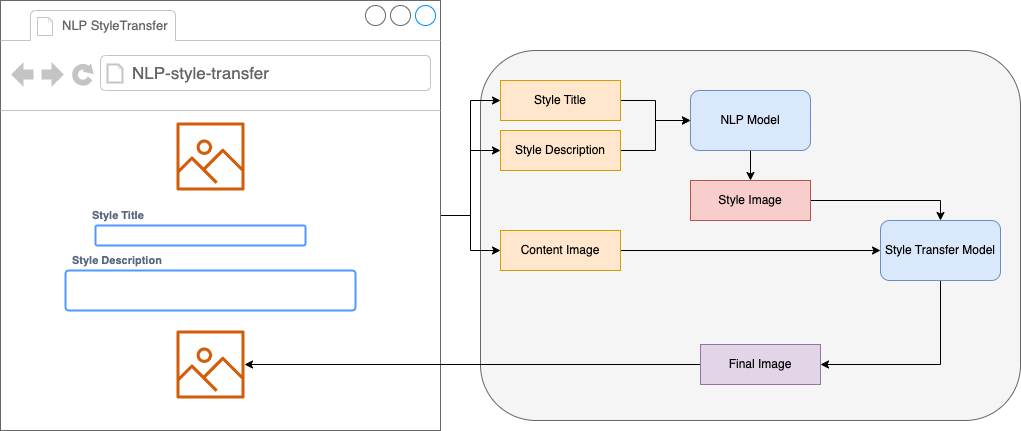}

   \caption{Overall Proof-Of-Concept Framework}
   \label{fig:pocframework}
\end{figure}

\begin{figure}[t]
  \centering
   \includegraphics[width=1\linewidth]{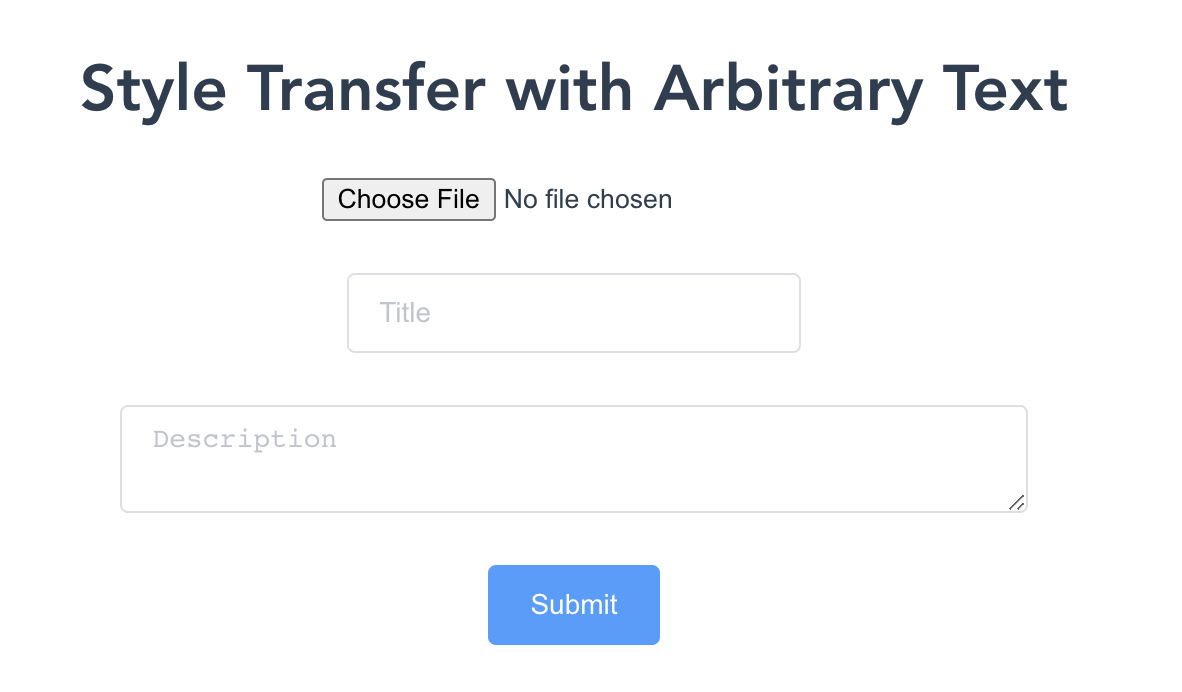}

   \caption{Tool User Interface}
   \label{fig:UI}
\end{figure}

To allow for easier testing and to bring the theory to life, we built and deployed an interactive website where can test our whole framework and model integration. The framework and pipeline connecting the models can be seen in Figure \ref{fig:pocframework}. The main technologies used include Vue.js and JavaScript libraries, supported with a Flask Python back end. The models were connected through a model handler function we implemented in the Flask back end. The input image and texts are taken from the front end and passed to the NLP model for embedding generation and finding the closest image to return. That image is saved and read in along with the input content image by the style transfer model for the purposes of generating the final image sent back to the front end. The proof-of-concept tool is intuitive and easy to use. A screenshot of the user interface is shown in Figure \ref{fig:UI}.\\

\begin{figure}[t]
  \centering
   \includegraphics[width=1\linewidth]{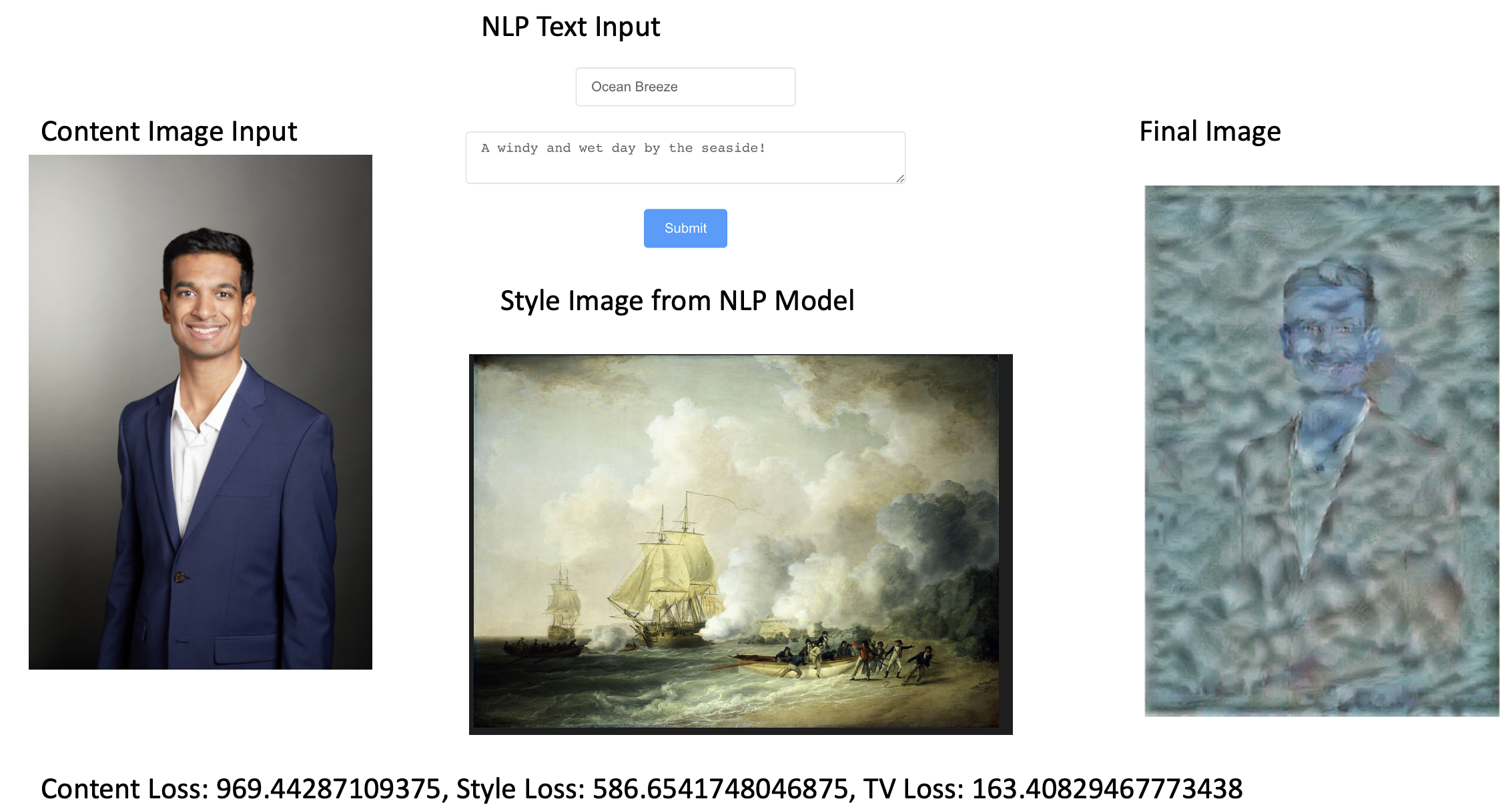}

   \caption{Example Proof-Of-Concept Output}
   \label{fig:pocoutput}
\end{figure}

\begin{figure}[t]
  \centering
   \includegraphics[width=1\linewidth]{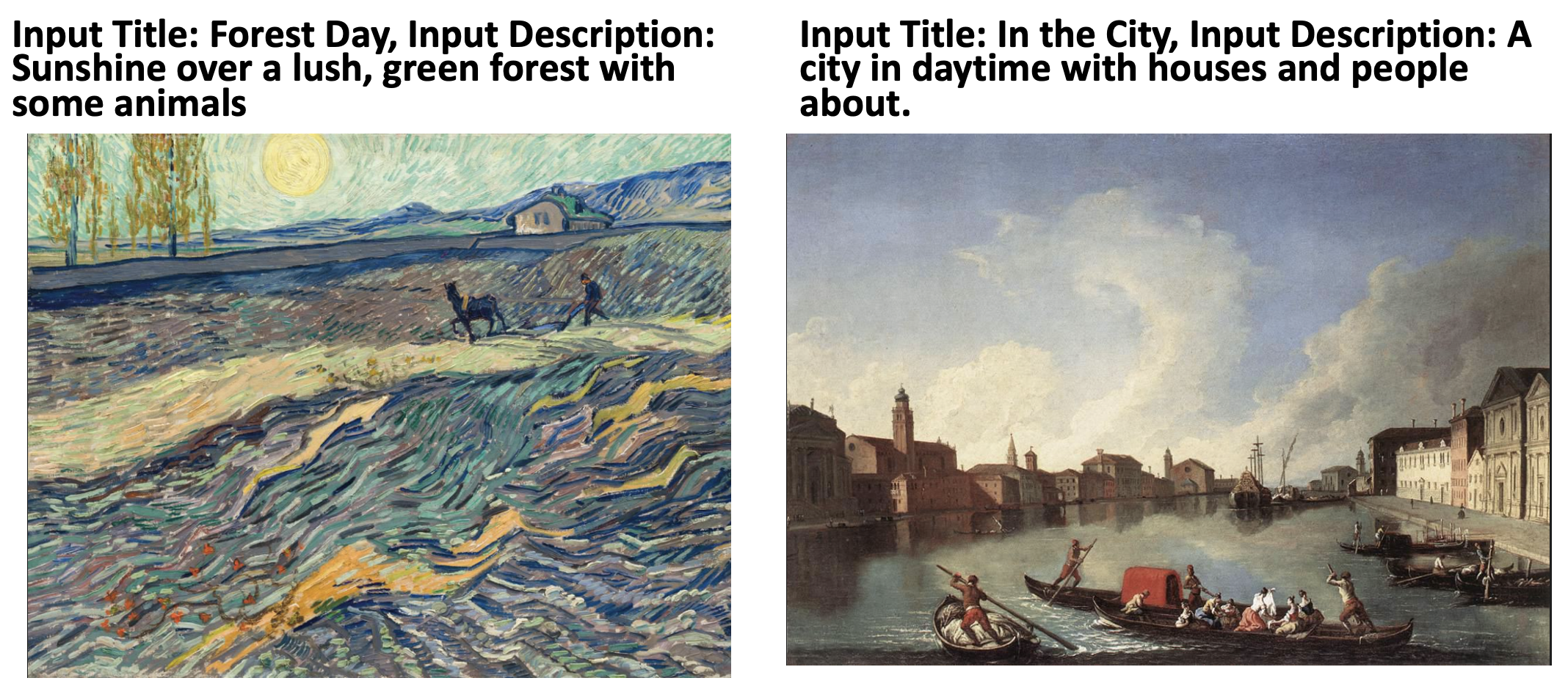}

   \caption{Example of Image Retrieval from Proof-Of-Concept Input}
   \label{fig:exTtoI}
\end{figure}

Upon entering the site, the user will be able to upload their local content image file using the select file button. The uploaded image will then be shown above the style title input area, and users can make sure that they upload the correct image. Then user can type in the style title and style description, clicking the button to submit when they are completed. Then, the front end will submit a request to the back end to process the style transfer task, after finishing the style transfer, it will return the final image to the front end. At last, the final image will be rendered by the front end, and the user can also use the original image shown in the front end for comparison. Example inputs and outputs to the proof-of concept tool integrating the two models, as well as loss values, can be seen in Figure \ref{fig:pocoutput}. Specific examples of the retrieved style image given a style title and style description input to the proof-of-concept tool can be seen in Figure \ref{fig:exTtoI}. A comparison to the baseline tests we ran can be seen in Table \ref{tab:finalcomp}. In this table, the title of the Style Image from Table \ref{tab:baseline} is used as the input text for the proof-of-concept tool to generate a final image and the resulting loss values.

\begin{table}
  \centering
  \begin{tabular}{cccc}
    \toprule
    Model & Content Loss & Style Loss & TV Loss \\
    \midrule
    SqueezeNet &2935.69 & 322.448 & 322.21 \\
    EfficientNet\_v2 & 23485.30 & 1710.36 & 43792.25 \\
    \bottomrule
  \end{tabular}
  \caption{Style Transfer Model Loss}
  \label{tab:results}
\end{table}

\begin{table}
  \centering
  \begin{tabular}{p{5cm}p{1cm}p{1cm}}
    \toprule
    Content Source, Style Title Text & Content Loss & Style Loss \\
    \midrule
    Tubingen, Composition VII & 1051.7 & 1427.7 \\
    Tubingen, Scream & 932.1 & 548.4 \\
    Tubingen, Starry Night & 284.8 & 328.9 \\
    Tubingen, Waterlilies & 398 & 278,8 \\
    Tubingen, Persistance of Memory & 859.8 & 252.5 \\
    Muse, Composition VII & 1610.7 & 323.4 \\
    Muse, Scream & 677.2 & 254.1 \\
    Muse, Starry Night & 194.4 & 497.9 \\
    \bottomrule
  \end{tabular}
  \caption{Example Results of Final Model}
  \label{tab:finalcomp}
\end{table}

\section{Conclusion}

Overall, this investigation helped to learn about how we can connect a language model to an image manipulation model and integrate something as concrete as English text with the abstractness of art. We were able to try and compare different NLP embedding approaches, as well as different techniques to preserve style when carrying out style transfer. Some experience in deployment of deep learning models was also gained through the creation of a proof-of-concept web tool. 

There are also opportunities to extend the work carried out in this report. The final version of the tool was not able to utilize transformer-based embedding through BERT with the same effectiveness. However, integrating a high-performing version of this is a possible area of future work. It would also be encouraged to try further model architectures for the style transfer model to see if any more advances in transfer quality can be gained. A possible follow-up work to this can also include the transfer of specific objects from image to image based on text, rather than just style. For example, if the content image was of a dog and the text was basketball, the final image would show a dog with a basketball. Another potential direction to go in would be to focus on creating outputs with strong artistic merit as a priority rather than loss functions. This slight change in objective might change how the models are created and executed.

Ultimately, this study was quite fruitful and successful in both its output and in the learning opportunities it created in terms of deriving style transfer from text.

{\small
\bibliographystyle{unsrt}
\bibliography{egbib}
}

\end{document}